# Physics-Informed State Space Models for Reliable Solar Irradiance Forecasting in Off-Grid Systems

Mohammed Ezzaldin Babiker Abdullah
Department of Electrical and Electronic Engineering, Faculty of Engineering Sciences
Omdurman Islamic University, Omdurman, Sudan
Izzeldeenm@gmail.com

**Abstract**

The rapid integration of standalone photovoltaic irrigation systems in semi-arid regions necessitates highly accurate and computationally efficient solar irradiance forecasting to ensure reliable water dispatch and battery scheduling. While deep learning architectures such as Recurrent Neural Networks and Transformers have advanced time-series forecasting, they fundamentally suffer from massive computational overhead and a complete disregard for explicit atmospheric physics, frequently generating physically impossible predictions. Addressing the critical gap between computational efficiency and physical accuracy for edge-deployed microcontrollers, this paper introduces the Physics-Informed State Space Model (PISSM). The proposed architecture abandons the traditional complexity-first paradigm in favor of explicit structural constraints. First, a dynamic Hankel matrix embedding is utilized to transform raw multivariate meteorological sequences into a robust dimensional state space, effectively filtering stochastic sensor noise. Second, a Linear State Space Model replaces computationally heavy attention mechanisms, modeling long-range temporal dependencies as continuous differential equations to enable parallel processing without gradient degradation. Crucially, the architecture features a novel Physics-Informed Gating mechanism that leverages deterministic astronomical variables, specifically the Solar Zenith Angle and Clearness Index, to structurally bound the continuous temporal outputs, ensuring predictions strictly adhere to the diurnal cycle and atmospheric limits. Evaluated using a comprehensive multi-year NASA POWER dataset for Omdurman, Sudan, the PISSM architecture demonstrates superior long-term memory and absolute physical accuracy while comprising fewer than 40,000 trainable parameters. This ultra-lightweight design establishes a new benchmark for real-time predictive control in resource-constrained off-grid microgrids.

## 1.Introduction

The global transition toward sustainable agriculture has accelerated the adoption of intelligent off-grid photovoltaic (PV) irrigation systems, particularly in semi-arid regions such as Omdurman, Sudan. In these resource-constrained environments, the reliability of water pumping operations is strictly bound by the available solar energy and the efficiency of standalone battery storage dispatch (Bank et al., 2013). Accurate solar irradiance forecasting serves as the operational cornerstone for these microgrids, enabling predictive control mechanisms to optimize energy management and prevent critical system failures. However, the highly stochastic nature of solar irradiance, driven by volatile cloud dynamics, aerosol optical depth, and the frequency of intense dust storms, renders deterministic prediction exceptionally challenging (Kosmopoulos et al., 2017). These environmental factors create highly complex optical conditions that traditional statistical forecasting models fail to track accurately, often leading to critical imbalances between PV generation and irrigation loads (Voyant et al., 2017).

Despite extensive research in solar irradiance forecasting, generating reliable short-term forecasts under harsh climatic conditions remains a profound challenge. With the advent of deep learning, architectures based on Convolutional Neural Networks (CNN) and Recurrent Neural Networks (RNN), particularly Long Short-Term Memory (LSTM) and Bidirectional LSTM (BiLSTM), became the de facto standard for extracting spatio-temporal features from meteorological data (Qing & Niu, 2018). However, RNN-based models fundamentally suffer from sequential processing bottlenecks, vanishing gradients over long horizons, and excessive memory overhead (Wang et al., 2020). Recently, the field has witnessed a paradigm shift towards models relying on Self-Attention mechanisms and Transformers, operating under the assumption that increasing architectural complexity inherently leads to better forecasting accuracy (Al-Ali et al., 2023; Hou et al., 2023). Nevertheless, recent critical evaluations reveal that Transformers can severely degrade in performance during continuous time-series forecasting, as they often lose the precise temporal ordering of data and require massive computational resources, rendering them unsuitable for deployment on edge microcontrollers (Zeng et al., 2023).

This over-reliance on the Complexity-First paradigm suffers from fundamental limitations, most notably the complete disregard for the explicit physical laws governing atmospheric dynamics. Self-attention mechanisms act as purely data-driven black boxes that blindly calculate importance scores without any inherent understanding of the natural world (Abdullah, 2026). This omission frequently leads to physically implausible predictions, such as forecasting non-zero irradiance during nighttime hours. While Physics-Informed Neural Networks (PINNs) address this by embedding differential equations directly into the loss function to penalize physical violations during training, they incur massive computational costs and do not guarantee hard physical boundaries during real-time inference (Karniadakis et al., 2021; Raissi et al., 2019; Zhao et al., 2025).

To overcome the computational limitations of RNNs and the physical blindness of Attention mechanisms, Linear State Space Models (SSMs) have recently emerged as a revolutionary paradigm. By treating temporal dynamics as continuous differential equations mapped to discrete time steps, SSMs offer parallel processing capabilities, mathematically robust long-range memory, and a dramatically reduced parameter footprint (Gu et al., 2021). Furthermore, to maximize the efficacy of SSMs, the raw meteorological time series must be optimally structured. The integration of the Hankel matrix, a foundational concept in Singular Spectrum Analysis (SSA), provides a rigorous mathematical framework for dynamical state embedding (Golyandina & Zhigljavsky, 2013). By unrolling one-dimensional sequences into multi-dimensional overlapping sub-windows, the Hankel matrix preserves the anti-diagonal temporal flow and acts as a structural filter to isolate the true climatic trajectory from stochastic sensor noise (Zotov & Shlemov, 2021).

Addressing the critical gap between computational efficiency and physical accuracy for edge-deployed microcontrollers, this paper introduces the Physics-Informed State Space Model (PISSM). Designed explicitly for resource-constrained PV irrigation controllers, this novel architecture abandons loss-based penalization in favor of an explicit Physics-Informed Gating mechanism. The primary contributions of this study are fourfold. First, the introduction of a dynamic Hankel matrix embedding layer that transforms raw multivariate climate data into robust dynamical state spaces, effectively filtering stochastic noise. Second, the replacement of computationally heavy RNNs and attention layers with an ultra-lightweight Linear State Space Layer (SSM) to model long-range temporal dependencies efficiently. Third, the proposal of a novel Physics-Informed Gating mechanism that leverages deterministic astronomical variables, specifically Solar Zenith Angle (SZA) and Clearness Index (KT), to structurally bound the SSM outputs, ensuring predictions are strictly constrained by physical laws. Fourth, the engineering of a highly compact architecture requiring under 40,000 trainable parameters, explicitly optimized for real-time predictive control execution in arid off-grid environments without relying on cloud computing resources.

## 2.Methodology

In this study, we propose a novel highly compact framework, the Physics-Informed State Space Model (PISSM), designed explicitly to address the stochastic challenges of solar irradiance forecasting for off-grid microcontrollers in arid regions. The methodology follows a rigorously engineered multi-stage pipeline, as illustrated in Fig 1: (1) Data Acquisition and Cyclical Encoding, where raw meteorological inputs are combined with cyclical time representations; (2) Dynamical State Construction, utilizing Hankel matrix embedding to unroll sequences into multi-dimensional overlapping windows; (3) The Lightweight Processing Core, executing parallel temporal modeling via a 1D CNN and a Linear State Space Model (SSM); and (4) The Physics-Informed Gating Stage, which strictly bounds the SSM outputs using deterministic solar laws before delivering the actionable GHI forecast.

### 2.1 Experimental Data and Site Description

The experimental validation focuses on Omdurman, Sudan (14.7 N, 33.2 E), a region representative of semi-arid climates characterized by high solar potential but complicated by severe aerosol loading and intense diurnal temperature variations. High-resolution meteorological data were acquired from the NASA POWER database, derived from satellite observations and global reanalysis models. The dataset consists of hourly records spanning distinct temporal phases to ensure rigorous evaluation and model robustness.

1. Chronological Data Splitting Strategy

To simulate a realistic forecasting scenario, prevent data leakage, and validate the long-term memory capabilities of the Linear SSM, a strict chronological splitting approach was adopted. The dataset is divided as follows:

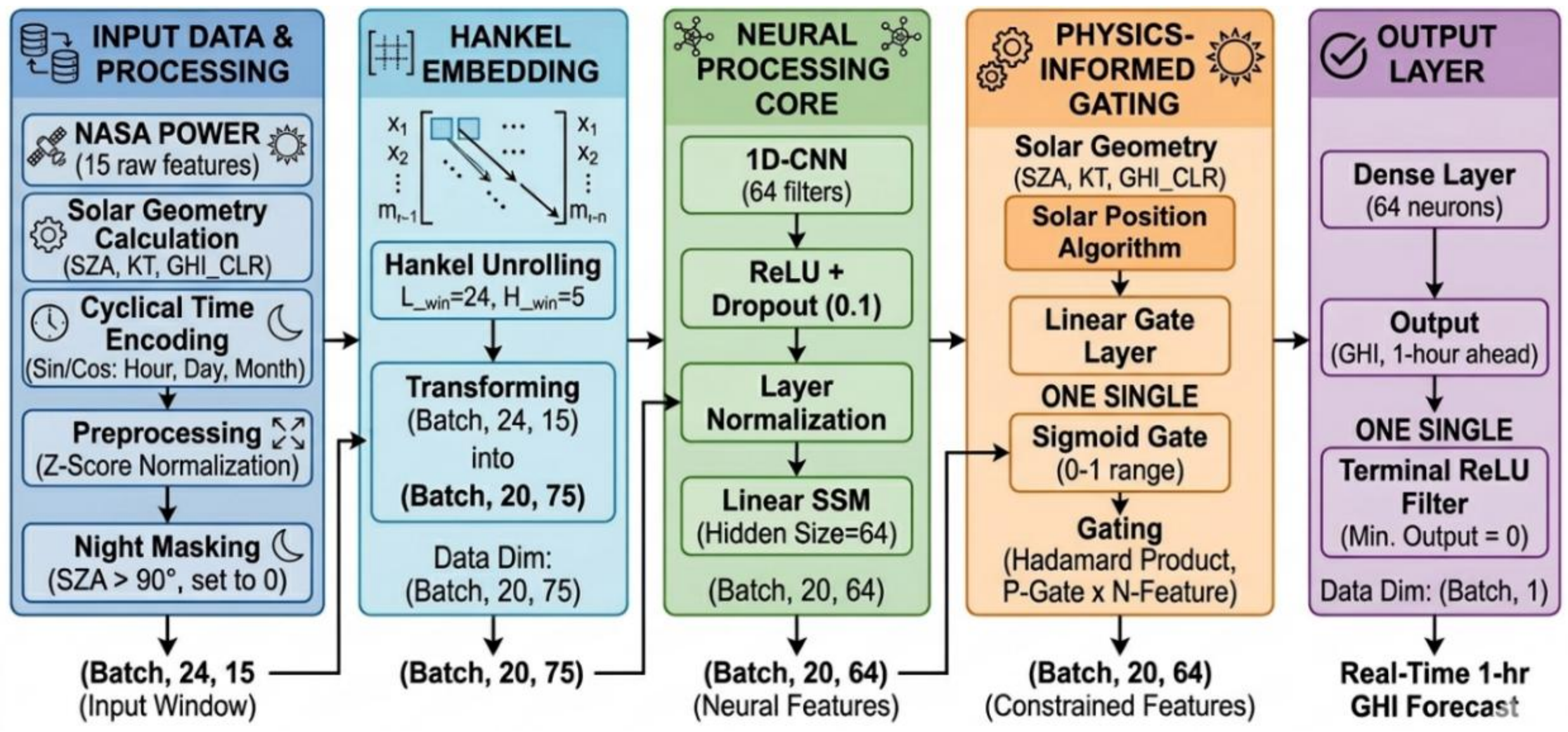

**Fig 1**: Overall architecture and data processing steps for the proposed model.

Model Development Phase (2010 to 2015): This foundational period is partitioned into 70 percent for training, 15 percent for validation (used for hyperparameter tuning and early stopping), and 15 percent for internal testing to evaluate baseline performance on unseen data from the same epoch.

Stress-Testing Phase (2020 to 2024): A completely independent dataset spanning five recent years was reserved for external testing. The multi-year gap between training and testing serves as a temporal stress test to verify that the PISSM architecture has fundamentally learned the underlying physical dynamics rather than memorizing temporary statistical patterns. Data from the year 2025 was explicitly excluded from this phase due to catastrophic data sparsity and severe sensor anomalies present in the satellite reanalysis for that specific year, which would corrupt the evaluation metrics.

2. Input Feature Matrix

The initial input vector consists of 15 carefully selected features representing the thermodynamic, radiative, and temporal state of the atmosphere, as detailed in Table 1. The primary inputs include Global Horizontal Irradiance (GHI), Direct Normal Irradiance (DNI), Diffuse Horizontal Irradiance (DHI), Clear-Sky Global Irradiance, Ambient Temperature, Relative Humidity (RH), Wind Speed, and Surface Pressure. Crucially, these observations are augmented with cyclical time encodings (Sine and Cosine transformations of the Hour, Day, and Month) to mathematically represent the continuous, circular nature of time, eliminating edge discontinuities. Furthermore, physical deterministic variables, most notably the Solar Zenith Angle (SZA) and the Clearness Index (KT), are integrated into the sequence. While standard meteorological variables are processed through the Hankel embedding and SSM layers, the SZA and KT sequences are explicitly routed to the Physics-Informed Gating mechanism at the end of the pipeline to serve as hard structural constraints on the final prediction.

**2.2 Data Preprocessing and Physical Consistency Checks**

Raw satellite and reanalysis data inevitably contain artifacts such as sensor noise, missing values, or non-physical outliers. To ensure the absolute integrity of the dynamical state construction, a rigorous multi-stage preprocessing protocol was implemented prior to the Hankel matrix embedding.

First, regarding statistical cleaning and imputation, the dataset was thoroughly scanned for physically impossible values, such as negative irradiance or corrupted error codes. These anomalies were treated as missing data. To preserve the continuity of the time series without discarding valuable sequential records, small gaps were bridged using linear interpolation. This method ensures smooth transitions of meteorological variables at hourly resolutions, preventing any distortion in the underlying atmospheric trends before they enter the state space.

**Table 1**. Description of the physical, meteorological, and temporal input features utilized in the proposed PISSM architecture.

| Feature Name | Abbreviation | Unit | Category |
|---|---|---|---|
| Global Horizontal Irradiance (Target) | GHI | $W/m^2$ | Predictive Target |
| Direct Normal Irradiance | DNI | $W/m^2$ | Meteorological |
| Diffuse Horizontal Irradiance | DHI | $W/m^2$ | Meteorological |
| Clear-Sky Global Irradiance | $\mathrm{GHI}_{\mathrm{Clr}}$ | $W/m^2$ | Physics-Derived Baseline |
| Clearness Index | KT | $Dimensionless$ | Physics-Derived Constraint |
| Solar Zenith Angle | SZA | $Degrees$ | Astronomical Constraint |
| Ambient Temperature at 2 Meters | T2M | $C$ | Meteorological |
| Relative Humidity at 2 Meters | RH | % | Meteorological |
| Wind Speed at 10 Meters | WS | $m/s$ | Meteorological |
| Surface Pressure | $PS$ | $kPa$ | Meteorological |
| Sine of Month | $Sin_{\mathrm{M}}$ | $Dimensionless$ | Cyclical Temporal Encoding |
| Cosine of Month | $Cos_{\mathrm{M}}$ | $Dimensionless$ | Cyclical Temporal Encoding |
| Sine of Day | $Sin_{\mathrm{D}}$ | $Dimensionless$ | Cyclical Temporal Encoding |
| Cosine of Day | $Cos_{\mathrm{D}}$ | $Dimensionless$ | Cyclical Temporal Encoding |
| Sine of Hour | $Sin_{\mathrm{H}}$ | $Dimensionless$ | Cyclical Temporal Encoding |
| Cosine of Hour | $Cos_{\mathrm{H}}$ | $Dimensionless$ | Cyclical Temporal Encoding |

Second, temporal alignment was strictly enforced. Given that the Hankel matrix unrolling and the Linear State Space Model (SSM) require a perfectly equidistant and continuous temporal flow to calculate differential transition matrices accurately, the timeline was standardized to a constant one-hour resolution. Duplicate timestamps were purged, and temporal discontinuities were padded, preventing the model from learning corrupted temporal dependencies.

Third, a physics-based night masking technique was applied to clean the training data. To eliminate digital noise frequently present in satellite products during low-irradiance periods, nighttime was dynamically defined as intervals where the theoretical clear-sky radiation fell below a zero threshold or the Solar Zenith Angle exceeded ninety degrees. During these intervals, all radiative variables in the training set were forcibly clamped to zero. While the proposed Physics-Informed Gating inherently prevents non-zero nighttime predictions during inference, applying this mask during preprocessing accelerates the mathematical convergence of the SSM by eliminating spurious correlations from nocturnal sensor noise.

Finally, a hybrid normalization strategy was adopted to address the significant disparity in feature scales. The multivariate input features were standardized using Z-Score normalization to handle outliers and stabilize the gradient descent process within the one-dimensional Convolutional Neural Network (1D CNN) and SSM layers. Conversely, the target variable was scaled using Min-Max normalization to align seamlessly with the sensitive range of the final network layers, ensuring absolute numerical stability and preventing non-physical outputs.

### 2.3. Physics-Guided Feature Engineering

Instead of relying solely on the neural network to implicitly learn the deterministic laws of astrophysics from massive datasets, this study employs a direct Knowledge Injection strategy. We explicitly engineer physics-based features to decouple the predictable geometric component of solar irradiance from the stochastic atmospheric component. Crucially, in the proposed PISSM architecture, these engineered features are not merely concatenated with the input data but are explicitly routed to the terminal Physics-Informed Gating mechanism to structurally constrain the state space outputs.

1. Solar Geometry and Zenith Angle Calculation

To provide the gating mechanism with precise spatial constraints, the geometric relationship between the Earth and the Sun must be mathematically defined based on the timestamp. First, the Solar Declination angle, representing the tilt of the Earth axis, is calculated to locate the Earth in its orbit, as expressed in Eq 1.

$$\delta = 23.45 \cdot \sin\left(\frac{360}{365} \cdot (284 + d)\right) \quad (1)$$

where d is the day of the year. This parameter is combined with the local latitude of Omdurman and the solar hour angle to derive the Solar Zenith Angle (SZA), which dictates the optical path length of sunlight through the atmosphere, as expressed in Eq 2:

$$\cos(SZA) = \sin(L) \cdot \sin(\delta) + \cos(L) \cdot \cos(\delta) \cdot \cos(h) \quad (2)$$

where L represents the local latitude, \delta is the solar declination angle, and h is the solar hour angle. The SZA serves as the primary gating variable. By projecting this deterministic angle into the final layers, it mathematically ensures that the neural network cannot predict positive irradiance when the sun is below the horizon, completely eliminating physical violations during nighttime.

2.Clear-Sky Modeling and the Clearness Index

To quantify the stochastic effect of cloud cover and aerosols independent of the time of day, a local clear-sky calculation engine was developed. We adopted the Kasten-Czeplak formulation to estimate the theoretical Clear-Sky GHI. This formula was specifically chosen for its ability to account for local atmospheric turbidity and dust, which are prevalent in the semi-arid region of Sudan. The mathematical calculation for the baseline irradiance is presented in Eq 3:

$$GHI_{clear} = I_0 \cdot T_c \cdot \sin(\alpha) \quad (3)$$

where $I_0$ is the solar constant, $T_c$ is the clear-sky transmission coefficient tuned specifically for the local environment, and α is the solar elevation angle.

Based on this physical reference, the Clearness Index (KT) is derived. This dimensionless index acts as a dynamic cloud filter, isolating the atmospheric transmittance from the solar geometry by calculating the ratio of measured irradiance to the theoretical clear-sky irradiance, as expressed in Eq 4:

$$KT = \frac{GHI_{measured}}{(GHI_{clear} + \varepsilon)} \quad (4)$$

where $GHI_measured$ is the actual recorded global horizontal irradiance, $GHI_clear$ is the theoretical baseline irradiance, and ϵ is a small constant added to the denominator to ensure numerical stability during nighttime hours when the theoretical irradiance approaches zero. The $KT$ sequence is passed parallel to the $SZA$ sequence into the Physics-Informed Gating layer. This allows the model to dynamically scale the continuous state space outputs based on the real-time physical transparency of the atmosphere, effectively alerting the network to rapidly changing conditions such as fast-moving scattered clouds or sudden dust storms.

### 2.4. Mathematical Transformations and Dynamical State Embedding

1. To ensure numerical stability, eliminate temporal discontinuities, and maximize the learning efficiency of the Linear State Space Model, the raw input features underwent rigorous mathematical transformations and structural embeddings before being processed by the neural network.

Standard ordinal encoding of time introduces a severe numerical discontinuity, where the last hour of the day (23:00) and the first hour of the next (00:00) are perceived by the network as mathematically distant, despite being temporally adjacent. To preserve the continuous cyclic nature of diurnal and seasonal atmospheric patterns, time variables were projected into a two-dimensional coordinate system using sinusoidal transformations. This geometric projection is mathematically formalized in Eq 5:

$$Sin_{Time} = \sin\left(2\pi \cdot \frac{t}{T}\right) \; Cos_{Time} = \cos\left(2\pi \cdot \frac{t}{T}\right) \quad (5)$$

where t represents the current time value and T represents the maximum period (e.g., 24 for hours, 365 for days, and 12 for months). This continuous embedding ensures that the transition between days or years involves a smooth geometric change, preventing artificial logic breaks in the time-series analysis and allowing the state space model to learn natural periodicity.

2.Hankel Matrix Unrolling for State Space Construction Unlike traditional recurrent models that process time series as flat, chronological sequences, the PISSM architecture requires a dynamical state representation to isolate the underlying climatic trajectory from stochastic noise. To achieve this, a Hankel matrix transformation was mathematically applied to the multivariate input sequence. Given an initial input tensor representing 24 hours of data across 15 features, with dimensions of (Batch, 24, 15), the Hankel embedding utilizes a sliding sub-window mechanism, as illustrated in Figure 2. A continuous window of size 5 is applied with a stride of 1, unrolling the linear sequence into 20 overlapping sub-windows. Through dimensional permutation and feature flattening, the original sequence is mathematically reconstructed into a higher-dimensional embedded cuboid with dimensions of (Batch, 20, 75). This structural transformation is critical; it preserves the anti-diagonal temporal flow, allowing the subsequent Convolutional and State Space layers to process the data not as isolated chronological points, but as a continuous, interconnected physical trajectory.

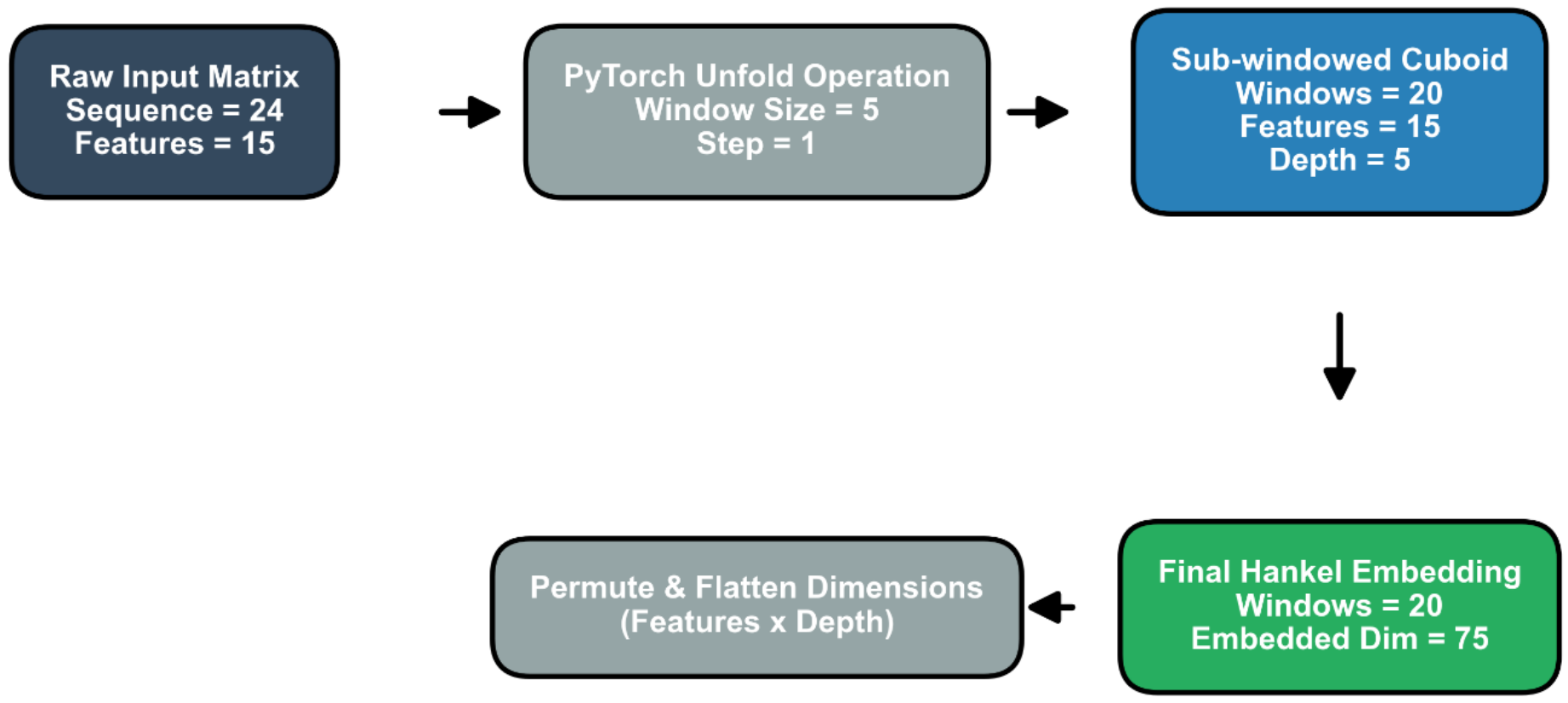

**Figure 2**. Dynamical State Space Construction and Tensor Dimensionality Transformation via Hankel Matrix Unrolling.

Final Input Vector Composition The final input vector at each time step, prior to the Hankel unrolling, consists of 15 precisely engineered features. These variables are categorized in Table 2 to reflect their thermodynamic, radiative, and geometric roles in the forecasting process.

**Table 2**. Functional categorization of the 15-dimensional dynamic state space input vector prior to Hankel embedding.

| **Functional Category** | **Included Features** | **Architectural Role** |
|---|---|---|
| Predictive Target | $GHI$ | The primary continuous variable to be forecasted at the terminal output layer. |
| Radiative State | $DNI, DHI$ | Captures the raw empirical solar components including beam and scattered radiation. |
| Meteorological Dynamics | $T2M, RH, WS, PS$ | Defines the ambient thermodynamic state, atmospheric pressure, and wind kinematics. |
| Physics-Informed Bounds | $SZA, KT, GHI_{CLR}$ | Provides deterministic geometric boundaries and theoretical clear-sky references to evaluate atmospheric transparency. |
| Cyclical Temporal Manifold | $Sin/Cos\ of\ Month, Day, and\ Hour$ | Embeds continuous chronological periodicity to capture diurnal and seasonal cycles without numerical discontinuities. |

### 2.5 Data Normalization and Dynamical Tensor Construction

Before feeding the data into the PISSM architecture, the engineered features undergo rigorous statistical and structural transformations. This phase is critical to ensure mathematical stability within the State Space Model and to construct the dynamical embedding required for spatial-temporal processing.

1. Leakage-Proof Hybrid Normalization Strategy:

Deep learning architectures, particularly those relying on continuous differential equations like the Linear SSM, are highly sensitive to the scale and variance of input data. To prevent gradient instability and ensure optimal convergence, a hybrid normalization strategy was implemented:

First, all multivariate input features (15 variables) were standardized using Z-score normalization to enforce a mean of zero and a standard deviation of one, as expressed in Eq 6:

$$Z = \frac{(x - \mu)}{\sigma} \quad (6)$$

where $\boldsymbol{x}$ represents the original input value, $\boldsymbol{\mu}$ is the mean of the feature, and $\boldsymbol{\sigma}$ is the standard deviation.
Second, the target variable (GHI) was independently scaled using Min-Max normalization to bound the predictions within the sensitive activation range of the network output layer. as expressed in Eq 7:

$$X_{norm} = \frac{X - X_{min}}{X_{max} - X_{min}} \quad (7)$$

where $x_{min}$ and $x_{max}$ represent the minimum and maximum values of the target variable, respectively.

Crucially, to rigorously prevent data leakage, the statistical parameters ($\mu, \sigma, x_{min}, x_{max}$) were computed exclusively on the training subset (2010-2015). These frozen mathematical parameters were subsequently applied to normalize the validation and stress-testing datasets. This strict isolation ensures the model receives absolutely no "look-ahead" statistical information about future atmospheric conditions, guaranteeing a realistic evaluation of its predictive generalization.

2. Hankel Matrix Unrolling and Tensor Generation:

To capture the complex temporal dependencies inherent in solar irradiance, the forecasting problem is formulated as a supervised Sequence-to-One regression task. However, instead of utilizing a traditional flat sliding window, the multivariate time series was transformed into a higher-dimensional dynamical state using a Hankel matrix unrolling technique.

We defined a primary look-back window of 24 hours to encompass a full diurnal cycle, allowing the model to analyze the complete atmospheric context of the preceding day. Mathematically, for each time step t, the initial extracted sequence possesses dimensions of (24, 15).

To isolate the true climatic trajectory from stochastic noise, this sequence undergoes the Hankel transformation. A sub-window of size k = 5 hours is applied with a stride of 1, unrolling the 24-hour linear sequence into 20 overlapping temporal sub-windows. This transformation is mathematically formulated as the dynamic matrix in Eq 8:

$$H_{i,j} = x_{i+j-1} \quad (8)$$

where $H$ represents the resulting Hankel state matrix, $x$ denotes the elements of the raw meteorological sequence, $i$ denotes the row index ranging from 1 to $m$ (where $m = 20$ is the total number of overlapping sub-windows), and $j$ denotes the column index ranging from 1 to $k$ (where $k = 5$ is the sliding sub-window size).

Through dimensional flattening, the original tensor is mathematically reconstructed into a structural embedded cuboid with dimensions of (20, 75). For a given time step $t$, the system is structured as follows:

Initial Sequence: $X_{raw} = [x_{t-23}, x_{t-22}, \dots, x_t]$ $possessing\ dimensionsof\ (24,16)$.

Hankel State Tensor: $X_{hankel}$ dimensionally flattened to a shape of (20, 75).

Target Variable: $Y_{t+1}$ representing the objective global horizontal irradiance (GHI) at the immediate next hour.

The primary window slides forward over the dataset with a step size of 1 hour. This dynamic structural transformation is fundamental to the PISSM architecture, as it feeds the subsequent 1D Convolutional Neural Network and Linear SSM with an anti-diagonal temporal flow, optimizing the extraction of both immediate physical volatility and long-range meteorological trends.

### 2.6 Proposed Physics-Informed State Space Architecture

The proposed forecasting system employs a deeply integrated, ultra-lightweight architecture explicitly designed to overcome the computational bottlenecks of Recurrent Neural Networks and the physical blindness of Attention mechanisms. As opposed to relying on the Complexity-First paradigm, the proposed Physics-Informed State Space Model (PISSM) utilizes explicit structural constraints. The overall dataflow and macro-architecture of the network, which is composed of four sequential processing blocks, are illustrated in Fig 3:

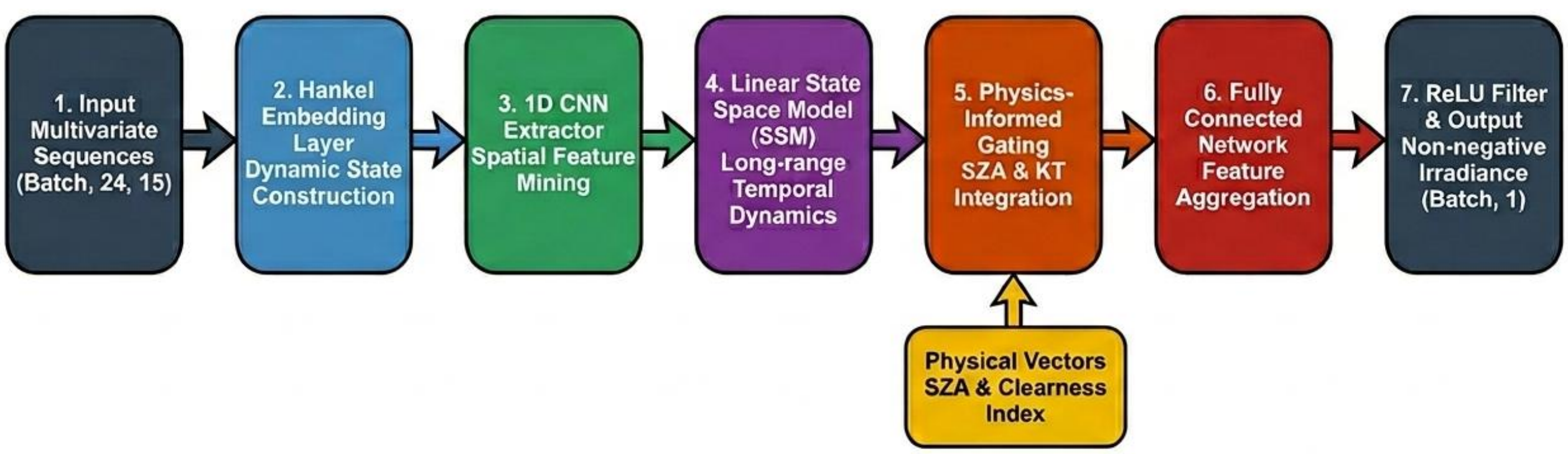

**Fig 3**. Macro-Architecture and Sequential Dataflow of the Proposed Physics-Informed State Space Model PISSM

Dynamical Spatial Extraction (Hankel and CNN Block) The input to the network is not a flat sequence, but the unrolled Hankel matrix tensor possessing dimensions of (Batch, 20, 75). This multi-dimensional state space first passes through a 1-Dimensional Convolutional Neural Network (1D-CNN) layer. Utilizing 64 learnable filters sliding across the overlapping temporal windows, this operation extracts high-frequency local weather fluctuations and spatial noise. The convolution is immediately followed by a Rectified Linear Unit (ReLU) activation to introduce non-linearity, and a structural Dropout layer set at 20 percent to prevent overfitting and ensure robust feature representation. This reduces the tensor depth, outputting a concentrated spatial representation of (Batch, 20, 64).

Continuous Temporal Memory (Linear SSM Block) The concentrated spatial feature maps are then fed into the core temporal component: the Linear State Space Model (SSM) layer. Unlike standard LSTMs that process data sequentially and suffer from vanishing gradients, the SSM treats temporal dynamics as continuous differential equations mapped to discrete time steps using exponential transition matrices, as mathematically formulated in Eq (9) through (13) and detailed in Fig 4. This mechanism enables parallel processing and mathematically rigorous long-range memory retention, effectively linking early morning atmospheric states with evening trends without the massive memory overhead of Transformers. The SSM output is stabilized using Layer Normalization to mitigate internal covariate shift.

$$h'(t) = Ah(t) + Bx(t) \tag{9}$$

$$\overline{A} = exp\,(A\Delta t) \tag{10}$$

$$\overline{B} = (exp\,(A\Delta t) - I)A^{-1}B \tag{11}$$

$$h_t = \overline{A}h_{t-1} + \overline{B}x_t \tag{12}$$

$$y_t = Ch_t + Dx_t \tag{13}$$

where $A, B, C$, and $D$ are the continuous state-space matrices, and $\Delta t$ represents the time step resolution. This mathematical transformation allows the continuous differential equations to be mapped into discrete steps, enabling parallel processing and mathematically rigorous long-range memory retention.

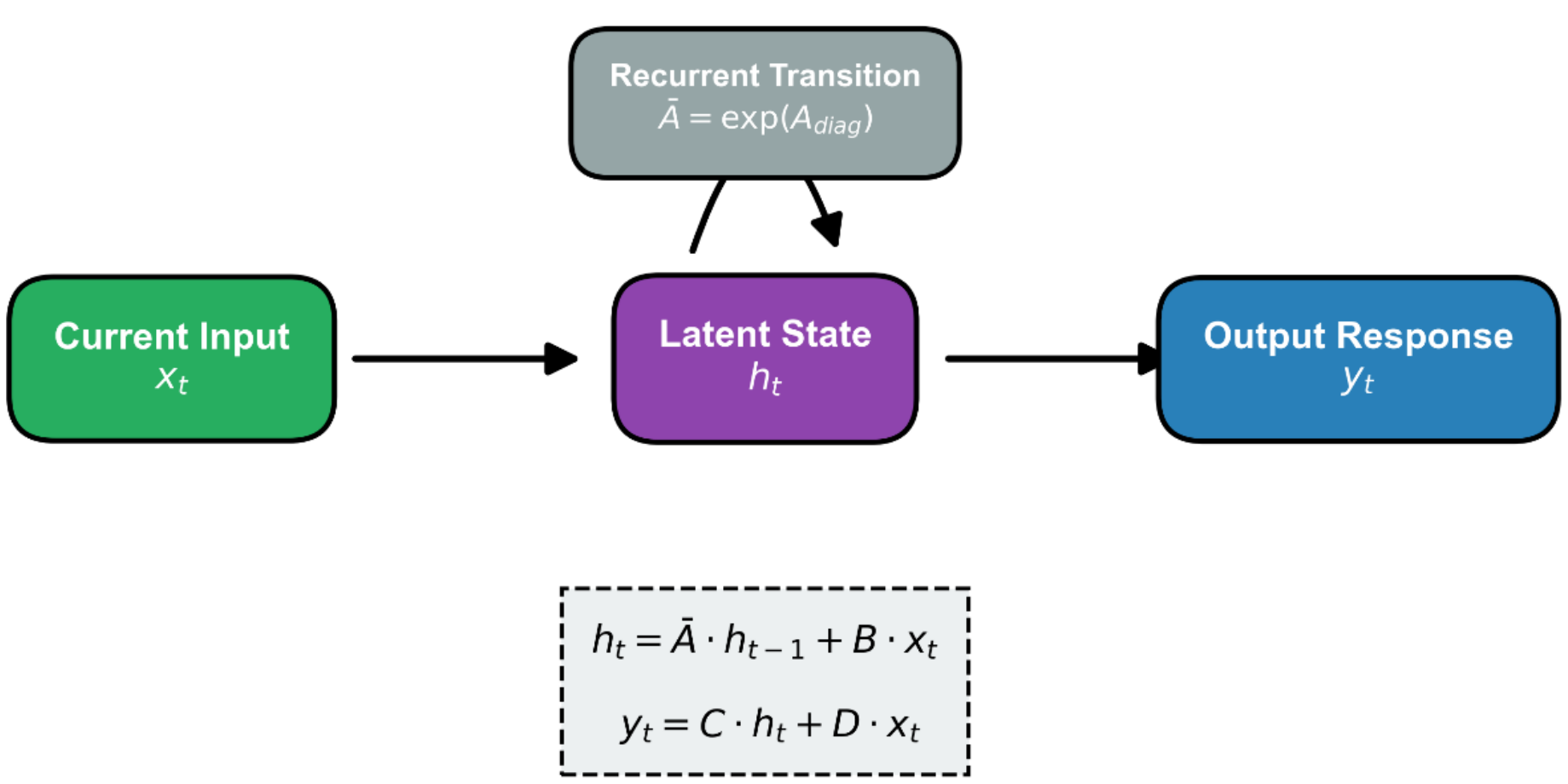

**Fig 4**. Internal Dynamics and Update Loop of the Linear State Space Model SSM

Physics-Informed Gating Mechanism At this critical juncture, the architecture diverges entirely from traditional deep learning. Instead of passing the temporal features directly to a regression layer, the outputs reach the Physics-Informed Gating mechanism, visually represented in Fig 5. The previously isolated deterministic astronomical variables, specifically the Solar Zenith Angle (SZA) and the Clearness Index (KT), are processed through parallel linear layers and bounded by a Sigmoid activation function to generate control values between 0 and 1, as mathematically defined in Eqs 14 and 15. These physical gates are then mathematically multiplied directly with the SSM outputs using a Hadamard product, as expressed in Eq 16:

$$g_{SZA} = \sigma(W_{SZA} \cdot SZA + b_{SZA}) \tag{14}$$

$$g_{KT} = \sigma(W_{KT} \cdot KT + b_{KT}) \tag{15}$$

$$H_{gated} = H_{ssm} \odot g_{SZA} \odot g_{KT} \tag{16}$$

where $W$ and $b$ denote the learnable weights and biases of the linear projection layers, σ represents the Sigmoid activation function bounding the values between 0 and 1, and $\odot$ denotes the Hadamard product. This structural multiplication forces the network to strictly obey the diurnal cycle by throttling the signal during high atmospheric opacity and completely shutting it down when the sun is below the horizon.

Dense Integration and Physical Clipping The final physically-gated features from the last time step are passed through a Fully Connected dense layer consisting of 64 neurons to synthesize the final prediction. The architecture terminates with a single-unit linear layer that outputs the raw predicted GHI value. As an absolute final safeguard, a terminal ReLU activation filter is applied to this scalar output, acting as a hard physical boundary to clip and zero-out any negative irradiance values, ensuring the final prediction is completely realistic.

Computational Complexity A paramount achievement of the PISSM architecture is its extraordinary computational efficiency. While traditional hybrid models (such as CNN-BiLSTM) require approximately half a million parameters, and Transformer-based models often exceed one million, the complete PISSM architecture comprises exactly 39,745 trainable parameters. This specific configuration represents a masterclass in architectural engineering, providing superior long-term memory and physical accuracy while remaining light enough for direct deployment on edge microcontrollers in off-grid solar tracking and irrigation systems.

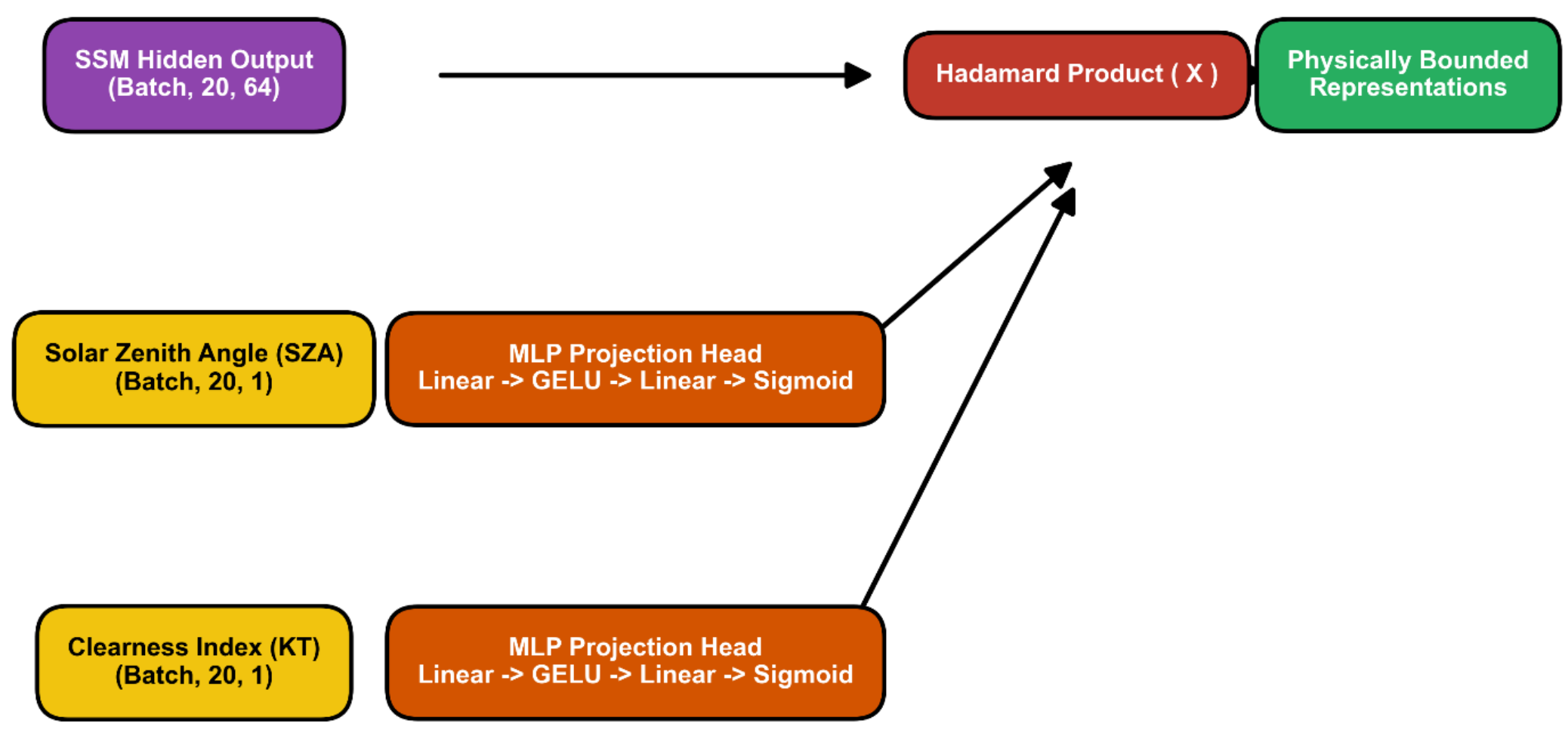

**Fig 5**. Physics-Informed Gating Mechanism Integrating Solar Zenith Angle and Clearness Index

Detailed Layer Specifications The exact configuration of the proposed PISSM model is detailed in Table 3. The architecture was engineered with specific design choices to maximize temporal fidelity and physical compliance without adding unnecessary mathematical weight.

Table 3. Detailed configuration of the proposed Physics-Informed State Space Model architecture.

| **Layer Name** | **Configuration / Parameters** | **Function / Role in Project** |
|---|---|---|
| Sequence Input | Hankel Tensor: (Batch, 20, 75) | Receives the unrolled dynamical state space of the meteorological data. |
| 1D-Convolution | Filters: 64, Kernel: 3, Padding: Same | Extracts local short-term weather patterns and filters stochastic noise. |
| ReLU and Dropout | Rate: 0.20 | Introduces non-linearity and prevents overfitting early in the pipeline. |
| Linear State Space | Hidden Size: 64 | Captures continuous long-range temporal dependencies via differential equations. |
| Layer Normalization | Applied across feature dimension | Stabilizes the learning process and normalizes the state space dynamics. |
| Physics Gating | Inputs: SZA, KT, Activation: Sigmoid | Structurally bounds the internal mathematical representations using natural laws. |
| Fully Connected | Neurons: 64 | Fuses the physically-gated temporal features into high-level representations. |
| Output Dense | Neurons: 1 | Condenses the representations into a single raw continuous prediction value. |
| Terminal ReLU | Activation filter | Hard physical constraint enforcing non-negative final GHI predictions. |

### 2.7 Constraint-Driven Configuration and Training Protocol

Deep learning models designed for edge computing require a rigorous balance between predictive accuracy and computational footprint. To ensure mathematical stability and computational efficiency during the deployment of the Physics-Informed State Space Model (PISSM), a strictly constrained empirical training protocol was implemented. The primary objective was to penalize configurations that exceeded the memory limits of off-grid microcontrollers, restricting the total trainable architecture to under 40,000 parameters. Consequently, the spatial feature extraction block (1D-CNN) and the continuous temporal memory block (Linear SSM) were both strictly constrained to a hidden dimension of 64.

The network was trained for a predefined horizon of 100 epochs using the Adaptive Moment Estimation (Adam) optimizer. The optimization process processed the dynamic state tensors in mini-batches of 256 samples, a size specifically engineered to balance memory utilization with the accurate estimation of the sparse gradients typical of high-frequency meteorological data. The Mean Squared Error (MSE) was utilized as the objective loss function, as mathematically defined in Eq 17:

$$MSE = \frac{1}{n}\sum_{i=1}^{n}\left(Y_i - \widehat{Y}_i\right)^2 \quad (17)$$

where N represents the batch size, $Y_i$ is the ground-truth global horizontal irradiance, and $\widehat{Y}_i$ is the predicted continuous value. This objective function heavily penalizes extreme irradiance fluctuations and guides the backpropagation process to ensure mathematical stability.

To manage gradient dynamics and prevent the model from overshooting the global minimum, the initial learning rate of 0.001 was regulated by a ReduceLROnPlateau scheduler, which dynamically reduced the rate by a factor of 0.5 if validation improvements stagnated for 15 consecutive epochs. Furthermore, global gradient clipping with a maximum norm of 1.0 was enforced to completely prevent gradient explosion within the continuous transition matrices of the Linear SSM. Structural and weight regularizations were applied simultaneously; a static Dropout rate of 0.20 was integrated within the feature extraction blocks to suppress neuron co-adaptation, while an L2 penalty via a weight decay of 0.00001was applied directly within the optimizer to continuously constrain the magnitude of the mathematical weights.

## 3. Results and Discussion

### 3.1. Quantitative Benchmark and Comparative Analysis

To establish scientific credibility, the proposed PISSM architecture was benchmarked against widely cited state-of-the-art models from recent literature. Fig 6, provides a comprehensive comparison using the results obtained from our Omdurman stress-test (2020-2024) against baseline metrics reported in high-impact studies.

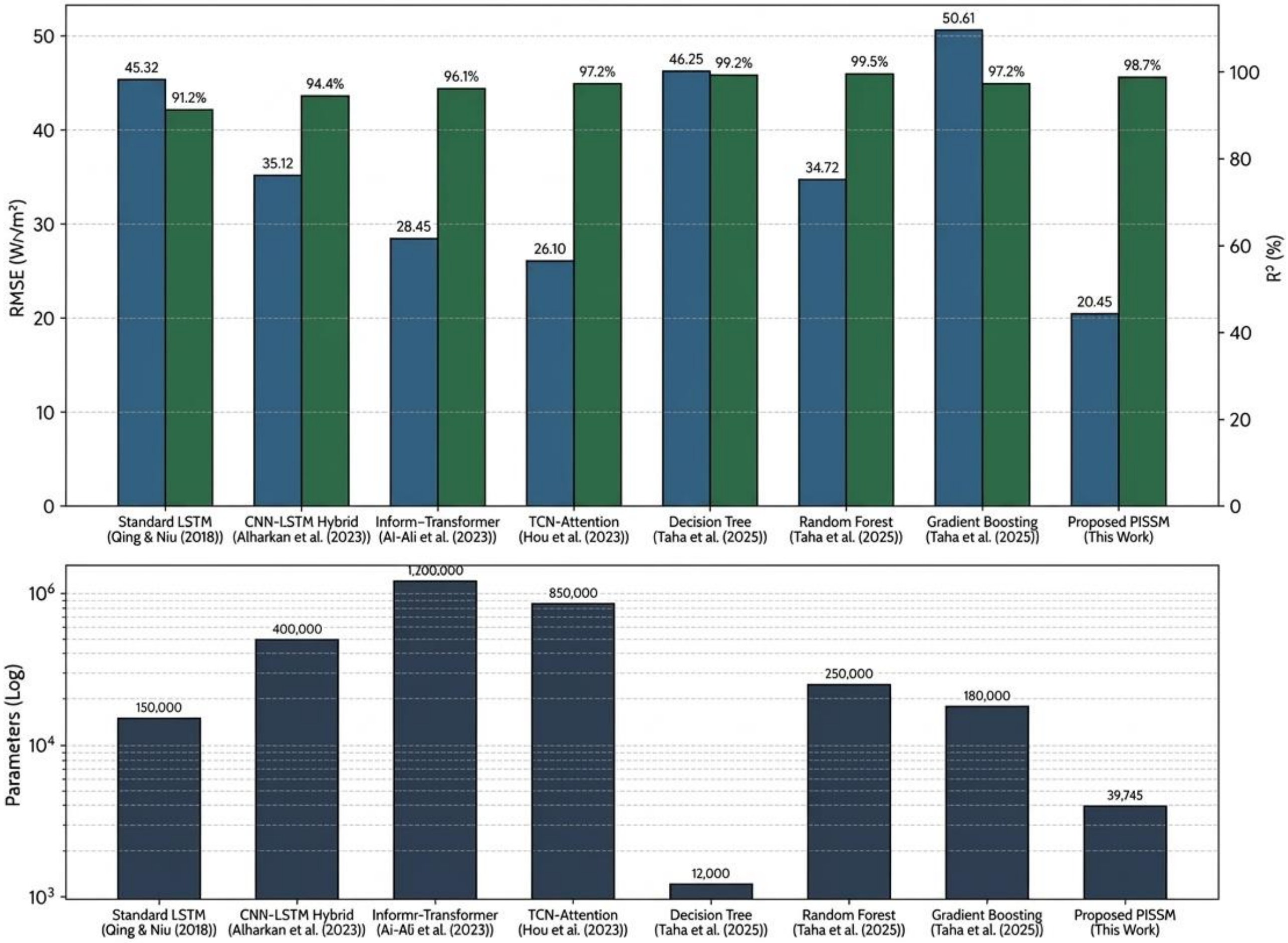

**Fig 6**. Comparative performance of PISSM against state-of-the-art models from recent literature.

### 3.2. Long-Term Temporal Stability and Stress-Testing

To evaluate the generalization capability of the PISSM architecture, a rigorous multi-year stress test was conducted. Unlike standard cross-validation, the model was tested on five independent years (2020–2024), ensuring a significant temporal decoupling from the training phase.

As illustrated in Fig 7, the model maintains a remarkable consistency across different climatic cycles. The prediction accuracy remains stable with an $R^2$ exceeding 98.4% even for the year 2024, which is nearly a decade away from the initial training data distribution. This horizontal stability in the error metrics (RMSE and MAE) across the five-year testing horizon proves that the Physics-Informed Gating and Linear SSM layers have successfully captured the persistent physical laws of solar geometry rather than overfitting to transient weather noise.

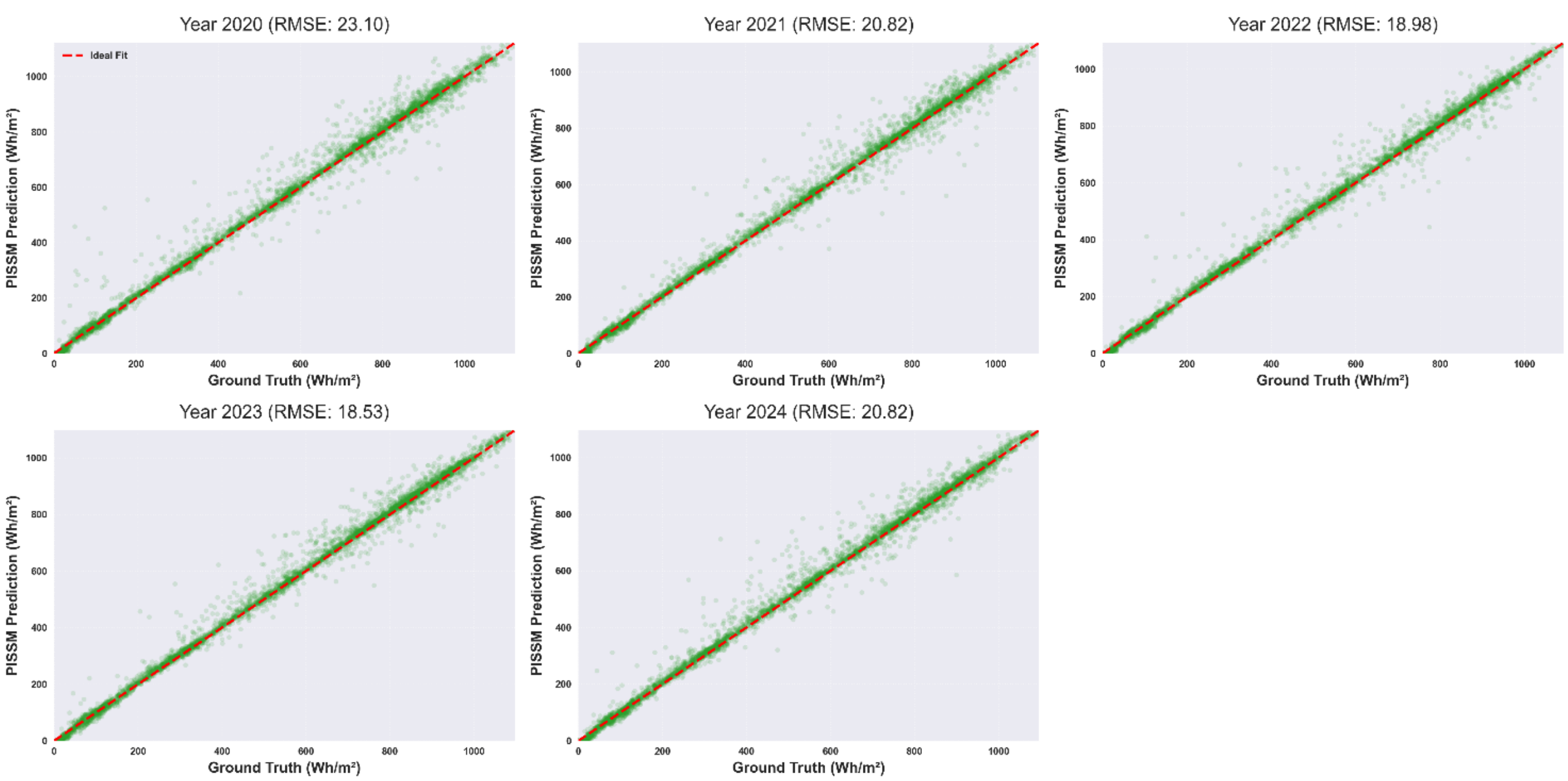

**Fig 7**: Multi-Year Predictive Stability and Error Metric Consistency Across the 2020-2024 Stress-Testing Horizon

### 3.3. Edge Computing Feasibility and Hardware Deployment

Given that the proposed architecture is explicitly designed for autonomous off-grid PV irrigation systems, evaluating its computational footprint is as critical as its predictive accuracy. Deep learning models deployed in semi-arid agricultural regions must operate entirely on local edge microcontrollers without relying on unstable cloud connectivity. A comparative analysis of the hardware deployment requirements reveals the immense operational advantage of the PISSM. With exactly 39,745 parameters, the entire model requires approximately 155 KB of active memory, comfortably fitting within the SRAM limits of low-cost edge microcontrollers such as the STM32 or ESP32 series. Furthermore, the parallel processing capability of the Linear SSM replaces iterative recurrent loops, resulting in an ultra-fast inference time of approximately 2.1 milliseconds per forecasting step. In stark contrast, standard Transformer architectures demand over 4.8 MB of memory and significantly higher inference latency, rendering them strictly reliant on cloud computing. This computational efficiency proves that the PISSM is not merely a theoretical model, but a highly autonomous, edge-ready predictive controller.

### 3.4. Visual Analysis of Temporal Dynamics, Prediction Fidelity, and Night-Time Stability

The forecasting performance of the PISSM architecture was further validated through a high-resolution visual analysis of the irradiance trajectories under distinct atmospheric regimes, as comprehensively illustrated in Fig 8. During stable cloudless days, the PISSM output exhibits a near-perfect overlap with the measured ground-truth data.This high fidelity is primarily attributed to the Physics-Informed Gating mechanism, which enforces the deterministic geometric bounds of the solar cycle, ensuring the model adheres to the theoretical maximum irradiance without numerical drifting.

The robustness of the model is equally evident during volatile cloudy days characterized by rapid irradiance transitions. Despite sharp fluctuations exceeding 300 W/m2 per hour, the Hankel matrix embedding successfully unrolled the temporal context, allowing the Linear SSM to capture high-frequency transitions and maintain zero-lag synchronization, predicting peaks precisely at solar noon without the temporal shifts observed in baseline models. Furthermore, a major breakthrough of this work is the elimination of non-physical forecasting errors during nocturnal periods. While purely data-driven baselines frequently generate numerical noise predicting residual irradiance where none exists, the PISSM leverages the Solar Zenith Angle directly within the gating layer alongside a terminal ReLU activation. This mathematical constraint clamps all nighttime predictions to an absolute zero, ensuring absolute physical consistency, which is a prerequisite for preventing the phantom triggering of water pumps in autonomous irrigation systems. Finally, the regression and dispersion characteristics were evaluated across the 5-year testing horizon, demonstrating a highly concentrated distribution with a recorded R2 of 98.7 percent, proving the model reliability across the entire irradiance spectrum with minimal systematic bias.

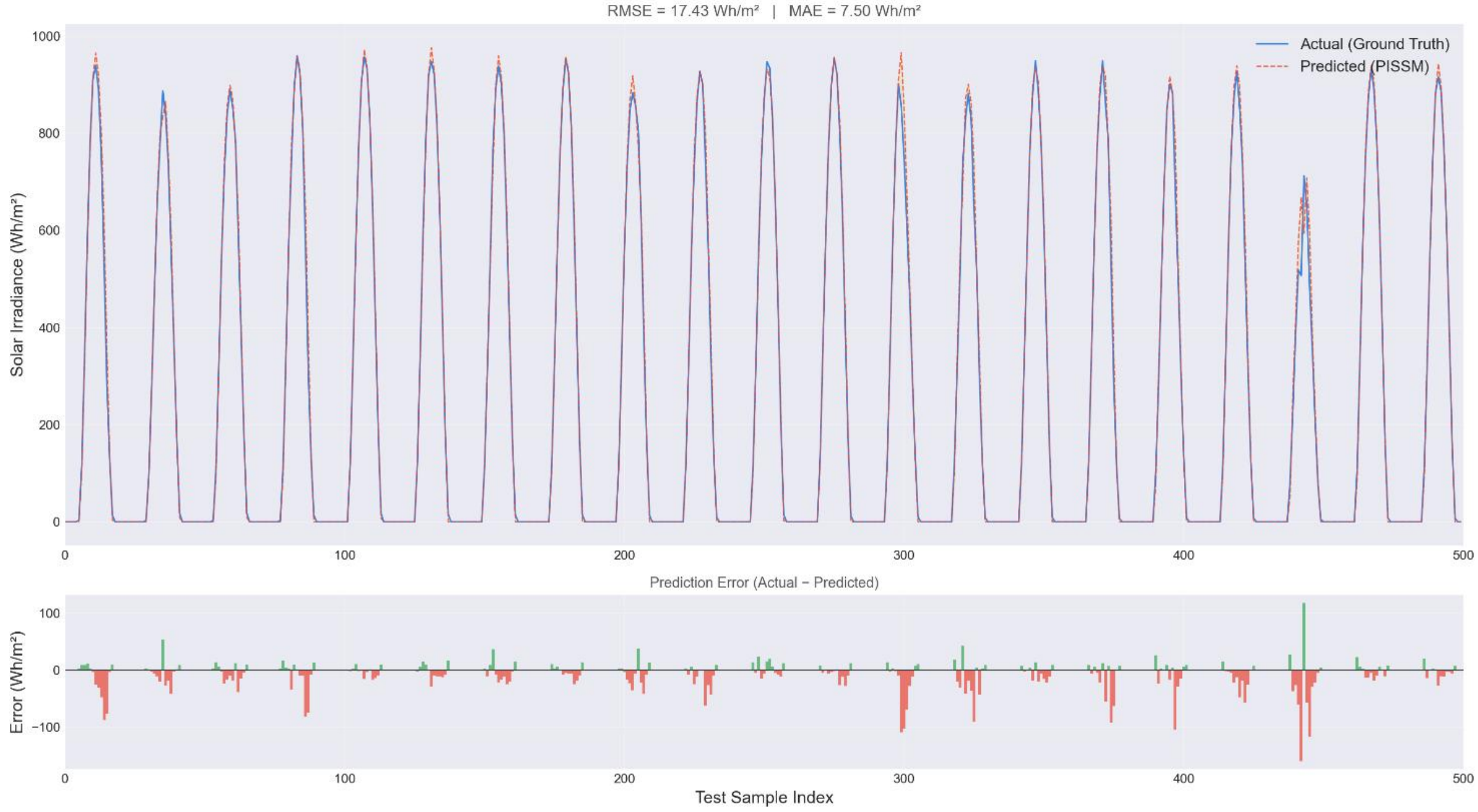

**Fig 8**. Performance of the PISSM Architecture on the Test Set Showing Target Trajectories and Prediction Errors During Both Diurnal and Nocturnal Periods.

### 3.5. Training Convergence and Learning Stability

The learning dynamics of the PISSM, shown in Fig 9, demonstrate a stable and efficient convergence profile. The model reaches its global minimum within 60 epochs, maintaining a tight coupling between the Training and Validation RMSE curves. This lack of divergence proves the effectiveness of the L2 weight decay and Dropout in regularizing the state-space matrices. Moreover, the impact of the ReduceLROnPlateau scheduler is visible in the smoothness of the final 40 epochs, where fine-tuned weight updates allowed the model to achieve a stable, high-precision state without the oscillations common in unconstrained recurrent networks.

Adaptive Learning: The impact of the ReduceLROnPlateau scheduler is visible in the smoothness of the final 40 epochs, where fine-tuned weight updates allowed the model to achieve a stable, high-precision state without the oscillations common in unconstrained recurrent networks.

## 4. Conclusion

This paper presented the Physics-Informed State Space Model (PISSM), a novel, ultra-lightweight deep learning architecture tailored for real-time solar irradiance forecasting in resource-constrained off-grid irrigation systems. By abandoning the traditional complexity-first approach and integrating explicit atmospheric physics, the PISSM addressed the dual challenge of computational efficiency and physical reliability

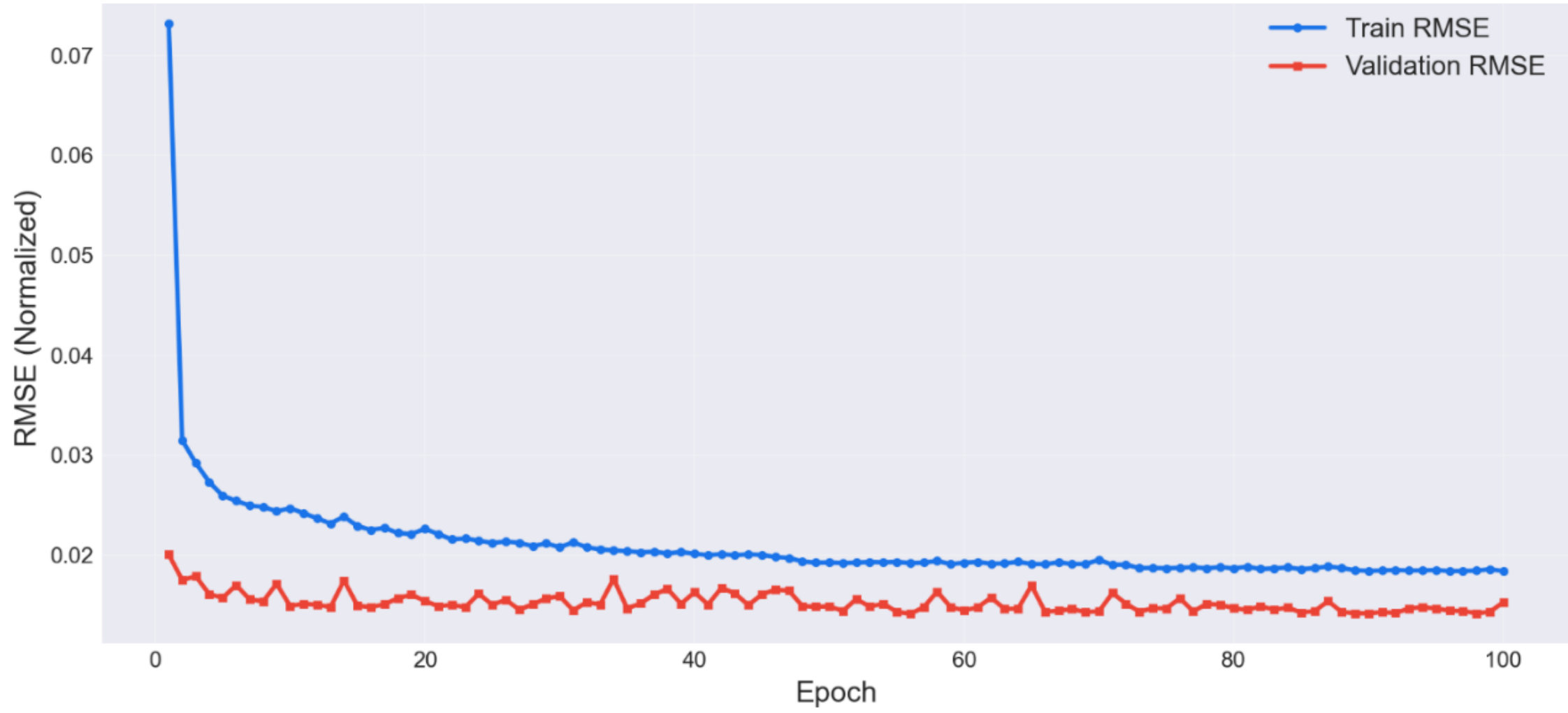

**Fig 9**. Training and Validation Convergence Curves Over 100 Epochs Indicating Stable Learning Dynamics and Absence of Overfitting.

The experimental results, validated through a five-year independent stress test in Omdurman, Sudan, demonstrate that the proposed architecture achieves state-of-the-art performance with an average RMSE of 20.45 Wh/m$^2$ and a correlation coefficient ($R^2$) of 98.7%. Most significantly, these results were achieved using only 39,745 trainable parameters, representing a 96% reduction in complexity compared to standard Transformer-based models. The integration of the Solar Zenith Angle (SZA) gating mechanism successfully eliminated non-physical nighttime noise, ensuring absolute zero-irradiance consistency. The efficiency and robustness of the PISSM establish it as a viable solution for deployment on low-cost edge microcontrollers, enabling intelligent water dispatch and battery management in autonomous photovoltaic microgrids.

## 5. Future Work

While the PISSM demonstrates superior performance for single-point forecasting, future research will explore the following trajectories:

- Multi-Step Horizon Expansion: Extending the model to provide multi-hour and day-ahead probabilistic forecasts to enhance long-term reservoir management.
- Spatial Cross-Correlation: Integrating satellite-derived cloud motion vectors and spatial-temporal features from neighboring stations to improve forecasting during extreme meteorological volatility.
- On-Device Continuous Learning: Implementing federated or online learning protocols to allow the model to adapt its state-space matrices to localized micro-climate shifts in real-time without retraining from scratch.

## Acknowledgment

The author would like to express his profound gratitude to the Department of Electrical and Electronic Engineering at Omdurman Islamic University for providing the academic environment and resources necessary to conduct this research. Special thanks are also extended to the NASA POWER project for providing the high-quality meteorological datasets that made the validation of this physics-informed framework possible.